\documentclass[conference]{IEEEtran}
\IEEEoverridecommandlockouts
\usepackage{cite}
\usepackage{amsmath,amssymb,amsfonts}
\usepackage{algorithmic}
\usepackage{graphicx}
\usepackage{textcomp}
\usepackage{xcolor}
\usepackage{epstopdf}
\usepackage{booktabs}
\usepackage{balance}
\usepackage{siunitx}
\usepackage{makecell}
\usepackage{tabularx}

\usepackage{fancyhdr}
\fancypagestyle{mahmood}{
  \fancyhf{} % clear all fields
  
  \fancyhead[C]{\footnotesize \textcopyright This work has been submitted to the IEEE for possible publication. Copyright may be transferred without notice, after which this version may no longer be accessible.}
}
\makeatletter
\let\ps@IEEEtitlepagestyle\ps@mahmood
\makeatother

\begin{document}

\title{Survey of Quantization Techniques for On-Device Vision-based Crack Detection\\
}

%I2MTC submission track
%Track Machine Learning and Big Data for Instrumentation and Measurement
\makeatletter
\newcommand{\linebreakand}{%
  \end{@IEEEauthorhalign}
  \hfill\mbox{}\par
  \mbox{}\hfill\begin{@IEEEauthorhalign}
}
\makeatother

\author{\IEEEauthorblockN{Yuxuan Zhang, Luciano Sebastian Martinez-Rau, Quynh Nguyen Phuong Vu, Bengt Oelmann and Sebastian Bader}
\IEEEauthorblockA{Department of Computer and Electrical Engineering \\ Mid Sweden University, Sundsvall, Sweden \\ yuxuan.zhang@miun.se} 
}
 
\maketitle

\begin{abstract}

Structural Health Monitoring (SHM) ensures the safety and longevity of infrastructure by enabling timely damage detection. Vision-based crack detection, combined with UAVs, addresses the limitations of traditional sensor-based SHM methods but requires the deployment of efficient deep learning models on resource-constrained devices. This study evaluates two lightweight convolutional neural network models, MobileNetV1x0.25 and MobileNetV2x0.5, across TensorFlow, PyTorch, and Open Neural Network Exchange platforms using three quantization techniques: dynamic quantization, post-training quantization (PTQ), and quantization-aware training (QAT). Results show that QAT consistently achieves near-floating-point accuracy, such as an F1-score of 0.8376 for MBNV2x0.5 with Torch-QAT, while maintaining efficient resource usage. PTQ significantly reduces memory and energy consumption but suffers from accuracy loss, particularly in TensorFlow. Dynamic quantization preserves accuracy but faces deployment challenges on PyTorch. By leveraging QAT, this work enables real-time, low-power crack detection on UAVs, enhancing safety, scalability, and cost-efficiency in SHM applications, while providing insights into balancing accuracy and efficiency across different platforms for autonomous inspections.
\end{abstract}


\section{Introduction}

Structural Health Monitoring (SHM) is essential for ensuring the safety and longevity of infrastructure such as bridges, buildings, and tunnels \cite{adin2023b,huang2024b}. By facilitating timely damage detection and assessment, SHM helps to prevent catastrophic failures, optimizes maintenance strategies, and reduces repair costs \cite{9335288,zhangtim2023a}. Traditional SHM systems use fixed sensors to monitor vibration, acoustic, or strain signals, offering insights into structural damage \cite{9964291,Ud2024}. However, these sensor-based approaches struggle to detect surface damage like cracks, which are critical early indicators of structural degradation. Vision-based crack detection has thus become an essential complement to SHM techniques. Advances in computer vision and deep learning have significantly improved the accuracy and scalability of automated crack detection compared to manual inspections \cite{7991459}. However, many existing studies focuses on the design of Convolutional Neural Networks (CNNs) and neglect the expense of the real-time deployability and energy-efficiency performance.

The adoption of unmanned aerial vehicles (UAVs) in SHM has garnered increasing attention \cite{10156415,10568977}. 
UAVs offer rapid and flexible capabilities for remote crack detection in complex environments, significantly enhancing vision-based inspection systems \cite{8453409}. 
Despite their potential, UAV applications are constrained by limited battery capacity, making energy efficiency a key challenge. 
Current UAV-based inspection systems rely on high-power platforms to run computationally intensive models, which not only reduce flight time but also decrease operational efficiency. 
For instance, Zakaria et al. in \cite{zakariaBridge2022a} evaluated solutions involving watt-level power platforms like Raspberry Pi, NVIDIA Jetson Nano, and Qualcomm Snapdragon 850 CPUs, but these platforms imposed significant constraints on flight time. 
Moreover, as model complexity increases, UAVs face greater challenges in real-time processing, further limiting their applicability in resource-constrained devices. Therefore, a low-power and resource-friendly solution is urgently needed to enable autonomous crack detection with UAVs.

The emergence of tiny machine learning (TinyML) offers promising solutions to these challenges \cite{Michele2023I2MTC,LucSAS2024,Michele2022TIM1,Michele2022TIM2}. 
TinyML enables machine learning inference on resource-constrained devices, reducing the need for computational resources and power consumption. 
This makes TinyML a highly efficient alternative for UAV-based visual inspections, extending flight time while achieving continuous, low-power crack detection and improving task efficiency. 
For example, \cite{zhangsas2024b} compared lightweight CNN models for crack detection using TinyML and found that MobileNet, despite being an earlier algorithm, demonstrated competitive performance in both lightweight design and detection accuracy. 
However, the usage of post-training quantization (PTQ) was found to reduce the detection accuracy by approximately 10\% in comparison to the original model. 
Alternatively, \cite{YuxuanZh2025} enhanced model performance by integrating image preprocessing techniques, at the cost of a significantly increased inference time (from 0.015 seconds to 4 seconds per inference). 
Thus, deploying efficient and accurate deep learning models on low-power platforms remains a challenge, particularly in balancing model complexity, preprocessing requirements, and detection accuracy.

Quantization techniques in TinyML have gained attention for their ability to reduce model size and computational requirements \cite{9869850}. 
By compressing model weights and activation functions from high-precision floating-point (FP) formats to 8-bit integer (INT8) representations, quantization enables hardware-friendly model deployment. 
The most common technique, PTQ, is easy to implement but can lead to significant accuracy loss, as demnstrated in \cite{zhangsas2024b}. 
In addition, existing research often focuses on single quantization techniques with specific deployment platforms, such as PTQ with TensorFlow (TF) Lite\footnote{https://ai.google.dev/edge/litert} or PyTorch\footnote{https://pytorch.org/} + Open Neural Network Exchange (ONNX)\footnote{https://onnx.ai/} for resource-constrained microcontroller (MCU) deployments. 
However, a systematic analysis of the performance differences across various training-deployment platforms is lacking. 
Additionally, studies on quantization techniques tailored for UAV-based on-device SHM applications are still lacking, leaving critical gaps in addressing the trade-off between accuracy and power consumption.

To address these gaps, this study systematically evaluates the performance of dynamic quantization, PTQ, and quantization-aware training (QAT) in a visual crack detection case study. 
We focus on the inference performance and trade-offs between different quantization techniques across various training and deployment platforms. 
Specifically, this work highlights the potential and limitations of quantization techniques for low-power platforms and proposes a practical TinyML optimization workflow. 
This workflow achieves milliwatt-level power consumption and millisecond-level on-device inference, supporting the efficient deployment of autonomous crack detection systems on UAVs. 
Moreover, this study provides scientific insights and design guidelines for the development and practical application of low-power SHM systems.

\section{Dataset}
This study utilizes the SDNET2018 dataset \cite{dorafshanSDNET2018AnnotatedImage2018}, a labeled collection of images commonly used as a benchmark for training, validating, and testing AI-based algorithms for concrete crack detection and classification. 
SDNET2018 contains over 56,000 images of concrete bridge decks, walls, and pavements, with and without cracks. 
A detailed breakdown of the dataset's composition is provided in Table \ref{tab:DatasetDescription}. 
The dataset includes images with various challenges, such as shadows, surface roughness, stains, edges, holes, and background debris. The images are in 256x256 JPEG RGB format, captured using a 16-megapixel Nikon camera at a working distance of 500 mm. Sample images from the dataset are shown in Figure \ref{fig:DataSamples}.

To ensure balance between samples with and without cracks, this study extracted 16,800 images from the original dataset. 
These were subsequently split into training, validation, and testing datasets in an 8:1:1 ratio. 
The final dataset consists of two categories (cracks and no racks), with 13,440, 1,680, and 1,680 images allocated to the training, validation, and test sets, respectively. 
All images were resized to 224x224 RGB format to match the input requirements of the pretrained networks.

\begin{table}[t]
\caption{SDNET2018 image dataset composition \cite{dorafshanSDNET2018AnnotatedImage2018}}
\begin{center}
\begin{tabular}{cccc}
\toprule
\textbf{\textit{Image description}}& \textbf{\textit{Cracks}} & \textbf{\textit{No cracks}}& \textbf{\textit{Total}}\\
\midrule
Reinforced   Bridge deck & 2025       & 11,595         & 13,620\\
Reinforced   Wall        & 3851       & 14,287         & 18,138\\
Unreinforced Pavement    & 2608       & 21,726         & 54,334\\
Total                    & 8484       & 47,608         & 56,092\\
\bottomrule
\end{tabular}
\label{tab:DatasetDescription}
\end{center}
\end{table}

%% FIGURE DATA DISTRIBUTION
\begin{figure}[tb]
      \centering
      \includegraphics[width=0.9\columnwidth]{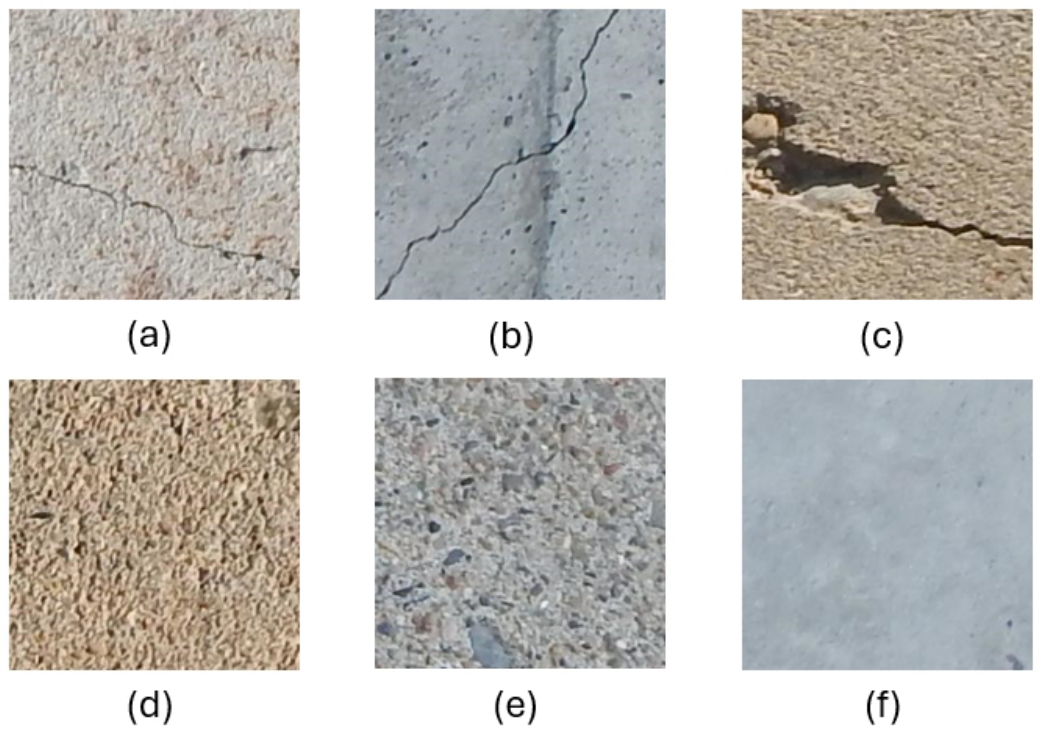}
      \caption{Image samples from SDNET2018, (a), (b), (c) with cracks, (d), (e), (f) without cracks.}
      \label{fig:DataSamples}
\end{figure}

\section{Methods}

This section introduces the selection of CNN models, quantization techniques, and TinyML toolchain used in this study. The overall workflow is illustrated in Fig.~\ref{fig:workflow}. 
Two models were trained separately using the TF and PyTorch frameworks. As shown in Fig.~\ref{fig:workflow}, these frameworks were combined with three different quantization techniques to generate \texttt{tflite} and \texttt{onnx} model files. Finally, the generated models were deployed on an MCU using STM32Cube.AI\footnote{https://stm32ai.st.com/stm32-cube-ai/}.

\begin{figure*}[tb]
      \centering
      \includegraphics[width=0.95\textwidth]{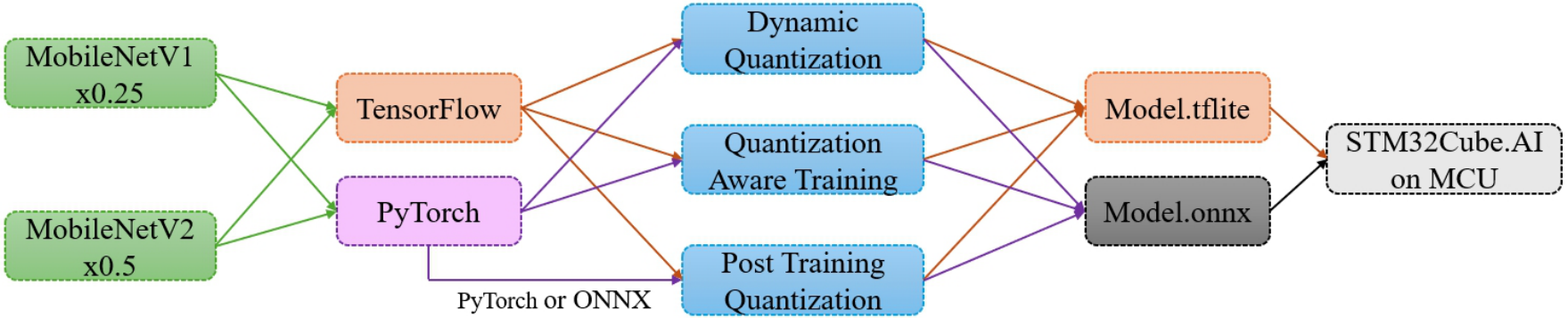}
      \caption{Workflow of quantization and deployment for MobileNet models on MCU.}
      \label{fig:workflow}
\end{figure*}

%有个表格， 以网络模型为最左行，或者流程图的形式
\subsection{Convolutional Neural Networks}
In \cite{zhangsas2024b}, the authors evaluated 14 CNN models with parameter counts of below 6 million parameters for visual crack detection. The study identified MobileNetV1 as the best-performing model. 
Building on this finding, this study employs two lightweight MobileNet versions: MobileNetV1 with width multiplier 0.25 (MBNV1x0.25) and MobileNetV2 with width multiplier 0.5 (MBNV2x0.5). 
MobileNetV1 leverages depthwise separable convolutions to reduce computations and model size while maintaining high accuracy. 
MobileNetV2 further enhances efficiency by introducing inverted residual blocks. 
The models used for TF training were sourced from Keras' pretrained models library\footnote{https://keras.io/api/applications/}, while PyTorch training utilized models from the open-source repository ImgClsMob\footnote{https://github.com/osmr/imgclsmob}. 
Both models were pretrained on the ImageNet-1K dataset. 
Notably, the \texttt{x0.25} version of MobileNetV2 was not selected due to limitations in the minimum width multiplier jointly supported by TF and ImgClsMob, which is \texttt{x0.5}.

Transfer learning and fine-tuning were applied to adapt the pretrained models to the SDNET2018 dataset for classification of cracks versus no cracks. 
During training, all layers were unfrozen to allow the weights of the entire network to be optimized. 
Additionally, the classification layers were modified to match the number of classes in the SDNET2018 dataset. 
This involved replacing the original fully connected layer with a new one, initialized with random weights and trained from scratch.

The training environment consisted of a Windows 10 64-bit operating system, an Intel\textregistered{} Core i9-12900 CPU, 32 GB of memory, and a single NVIDIA RTX 3090 GPU with 24 GB of memory. 
Training was performed using PyTorch 2.5.1 and TF 2.8.0 frameworks. 
All models were fine-tuned for 10 epochs with a learning rate of 0.001.

To evaluate the classification performance of the models, the F1-score was chosen as a key metric. 
The F1-score provides a balanced measure of precision and recall in a single value, ensuring that the model’s ability to accurately identify true positives is not overshadowed by the presence of false positives and false negatives. 
Specifically, $TP_{i}$ represents the count of correctly classified instances for class~$i$, $FP_{i}$ refers to the instances incorrectly identified as class~$i$ (false positives), and $FN_{i}$ represents the ground truth instances of class~$i$ that were not recognized (false negatives). 
A micro-averaged F1-score was computed, aggregating false positives, false negatives, and true positives across both classes. The F1-score for each class~$i$ (cracks and non-cracks) was calculated as follows:
\begin{equation}
F1\text{-}score_{i} = \frac{2 \cdot TP_{i}}{2 \cdot TP_{i} + FP_{i} + FN_{i}}.
\end{equation}

\subsection{Quantization Techniques}
To optimize the deep learning models for deployment on resource-constrained devices, we explored three common quantization techniques: Dynamic Quantization, PTQ, and QAT. These approaches aim to reduce model size, improve inference efficiency, and minimize energy consumption while maintaining acceptable performance levels.

\textbf{Dynamic Quantization:} Dynamic quantization reduces the precision of weights and activations (e.g., to INT8) dynamically during inference, based on their runtime range. This approach requires no modifications to the model during training or post-processing, making it computationally efficient. However, it may result in minor accuracy loss due to the lack of fine-tuning for quantization-induced errors.

\textbf{Post-Training Quantization:} PTQ converts a pre-trained model from full precision (e.g., FP32) to lower precision (e.g., INT8) using a calibration dataset. It is simple to implement and requires no additional training, but its effectiveness depends on the quality of the calibration dataset. PTQ is well-suited for resource-constrained devices.

\textbf{Quantization-Aware Training:} QAT incorporates quantization effects during training to enable the model to adapt to lower-precision constraints. This technique requires additional training resources but typically achieves better accuracy than PTQ or dynamic quantization, particularly for complex tasks.

These three techniques were evaluated on the target dataset and hardware platform, with performance measured in terms of accuracy, inference time, memory consumption, and computational efficiency. Detailed experimental results are provided in the subsequent sections.

\subsection{Tiny Machine Learning}

The details of the development board and the target MCU used in this study are summarized in Table~\ref{tab:BoardDetails}. The TinyML toolchain used in this study, as illustrated in Fig.~\ref{fig:workflow}, consists of three distinct branches:

\textbf{TensorFlow-based Workflow:} In this branch, CNN models trained using TF are quantized using three quantization techniques and converted into \texttt{tflite} format. These models are then processed using STM32Cube.AI to generate C/C++ code for deployment on low-power, resource-constrained MCUs.

\textbf{PyTorch-based Workflow:} In this branch, CNN models trained using PyTorch are quantized using three quantization techniques and converted into \texttt{onnx} format. The C/C++ code generation remains the same as in TF-based workflow. 

\textbf{ONNX-based Workflow:} In this branch, FP models trained using PyTorch are first converted into \texttt{onnx} format. PTQ is then performed using ONNX. The quantized models' \texttt{onnx} file is converted into C/C++ as in the TF-based workflow. Other quantization techniques are not used in this branch due to lack of support in ONNX (version 1.18.0).

\begin{table}[t]
\caption{Technical specifications of STM32H7B3I-DK Discovery Kit}
\begin{center}
\begin{tabular}{cc}
\toprule
\textbf{\textit{Parameter}}& \textbf{\textit{STM32H7B3I-DK Discovery Kit}} \\
\midrule
MCU                 &  STM32H7B3LIH6Q  \\
CPU Core            &  ARM Cortex M7 (32-bit)\\
CPU Frequency       &  280MHz   \\
SRAM                &  1.4MB    \\
Flash               &  2MB   \\
FPU                 &  \checkmark \\
Voltage   &  3.3V   \\

\bottomrule
\end{tabular}
\label{tab:BoardDetails}
\end{center}
\end{table}

\section{Results and Discussion}
This section presents the performance impact of different quantization techniques on vision-based on-device crack detection tasks across various design branches.

\subsection{Model Performance}
In this section, we analyze the performance of the two models, MBNV1x0.25 and MBNV2x0.5, across different quantization techniques and platforms. The evaluation is based on their F1-scores for two binary classification tasks: detecting the presence or absence of cracks.

\textbf{MBNV1x0.25:} For MBNV1x0.25, the FP models trained in TF and PyTorch serve as the baseline, achieving F1-scores of 0.8252 and 0.8485 for the crack task, and 0.8318 and 0.8479 for the non-crack task, respectively. The results are shown in Fig. \ref{fig:moibilenetv1}, highlighting PyTorch's slightly superior performance. Dynamic quantization has minimal impact on the performance, with F1-scores close to the FP models in both tasks. PTQ results in noticeable accuracy degradation, with TF-PTQ, Torch-PTQ, and ONNX-PTQ achieving 0.7262, 0.8057, and 0.7895 for the crack task, and 0.6424, 0.7850, and 0.7471 for the non-crack task, respectively. While ONNX-PTQ balances accuracy and resource efficiency, TF-PTQ exhibits significant accuracy loss. QAT mitigates this degradation, achieving F1-scores of 0.8267 (TF) and 0.8209 (PyTorch) for the crack task, and 0.8145 (TF) and 0.8020 (PyTorch) for the non-crack task, demonstrating robustness across both classifications.

\textbf{MBNV2x0.5:} The performance results for MBNV2x0.5 are presented in Fig. \ref{fig:moibilenetv2}. FP models set the baseline, with TF-Float and Torch-Float achieving F1-scores of 0.8240 and 0.8256 for the crack task, and 0.7762 and 0.8068 for the non-crack task. Dynamic quantization slightly improves performance for the crack task, with F1-scores of 0.8309 (TF) and 0.8256 (PyTorch), and shows no loss for the non-crack task. PTQ reveals larger accuracy gaps, with TF-PTQ, Torch-PTQ, and ONNX-PTQ scoring 0.7868, 0.7282, and 0.7839 for the crack task, and 0.6722, 0.5692, and 0.6967 for the non-crack task, respectively, indicating PTQ's limitations, particularly in Torch's non-crack classification. QAT achieves performance close to FP models, with F1-scores of 0.8285 (TF) and 0.8463 (PyTorch) for the crack task, and 0.7990 (TF) and 0.8289 (PyTorch) for the non-crack task, demonstrating its ability to balance accuracy and efficiency.

Across both models, PyTorch implementations generally achieve higher F1-scores than their TF counterparts in FP and dynamic quantization scenarios. Dynamic quantization consistently preserves FP accuracy, making it a practical choice for models where retraining is not feasible. PTQ, while resource-efficient, introduces noticeable accuracy drops, particularly in TF-based implementations for non-crack task. ONNX-PTQ offers a balanced approach but does not consistently outperform platform-specific PTQ techniques. QAT stands out as the most robust quantization technique, achieving F1-scores comparable to FP models across both classification tasks, albeit at the cost of additional training time.
This analysis provides insights into the trade-offs between different quantization techniques and highlights the strengths of QAT in preserving model performance.

\begin{figure}[tb]
      \centering
      \includegraphics[width=0.95\columnwidth]{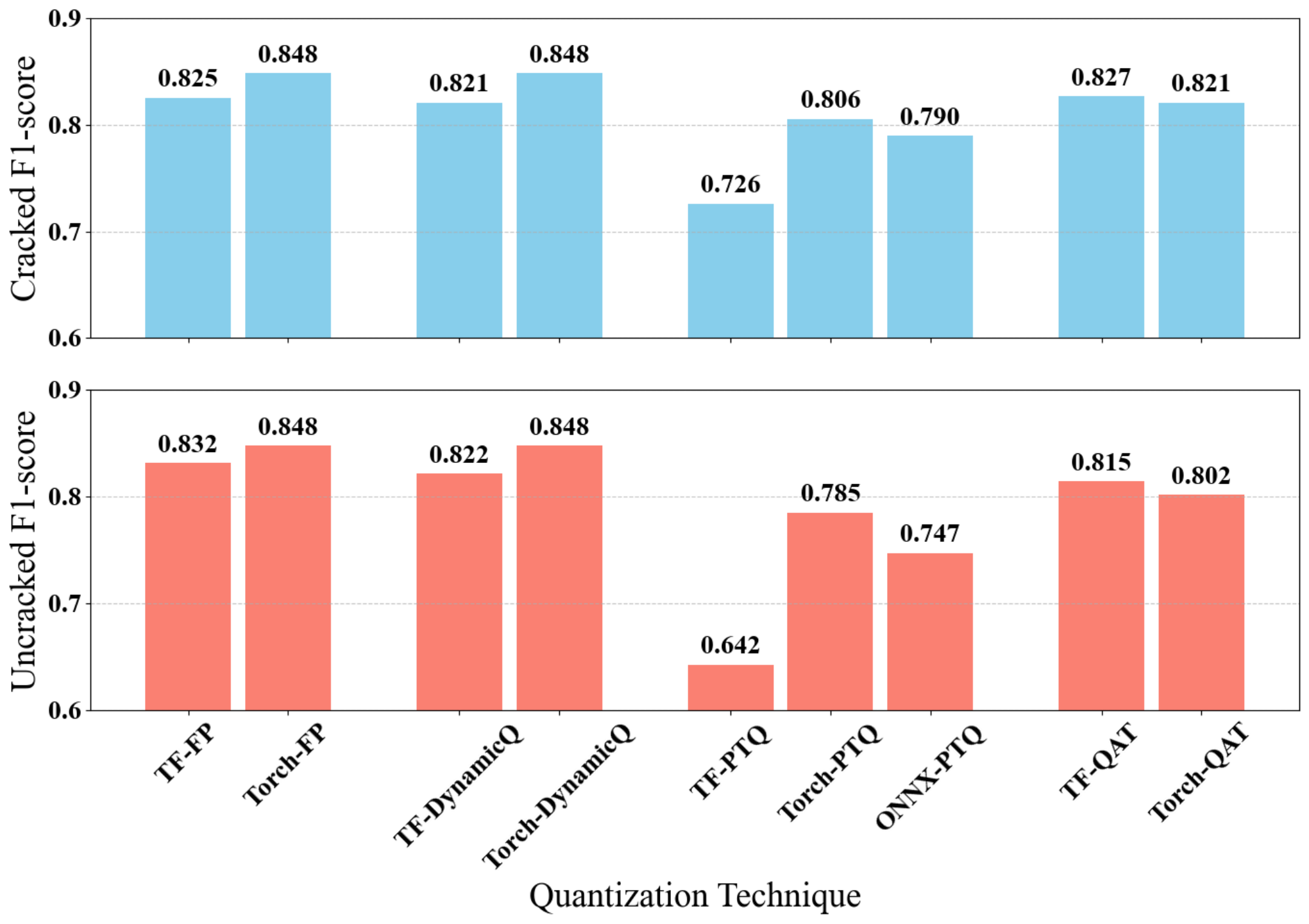}
      \caption{Performance of different quantization techniques with TF and Pytorch on MBNV1x0.25.}
      \label{fig:moibilenetv1}
\end{figure}

\begin{figure}[tb]
      \centering
      \includegraphics[width=0.95\columnwidth]{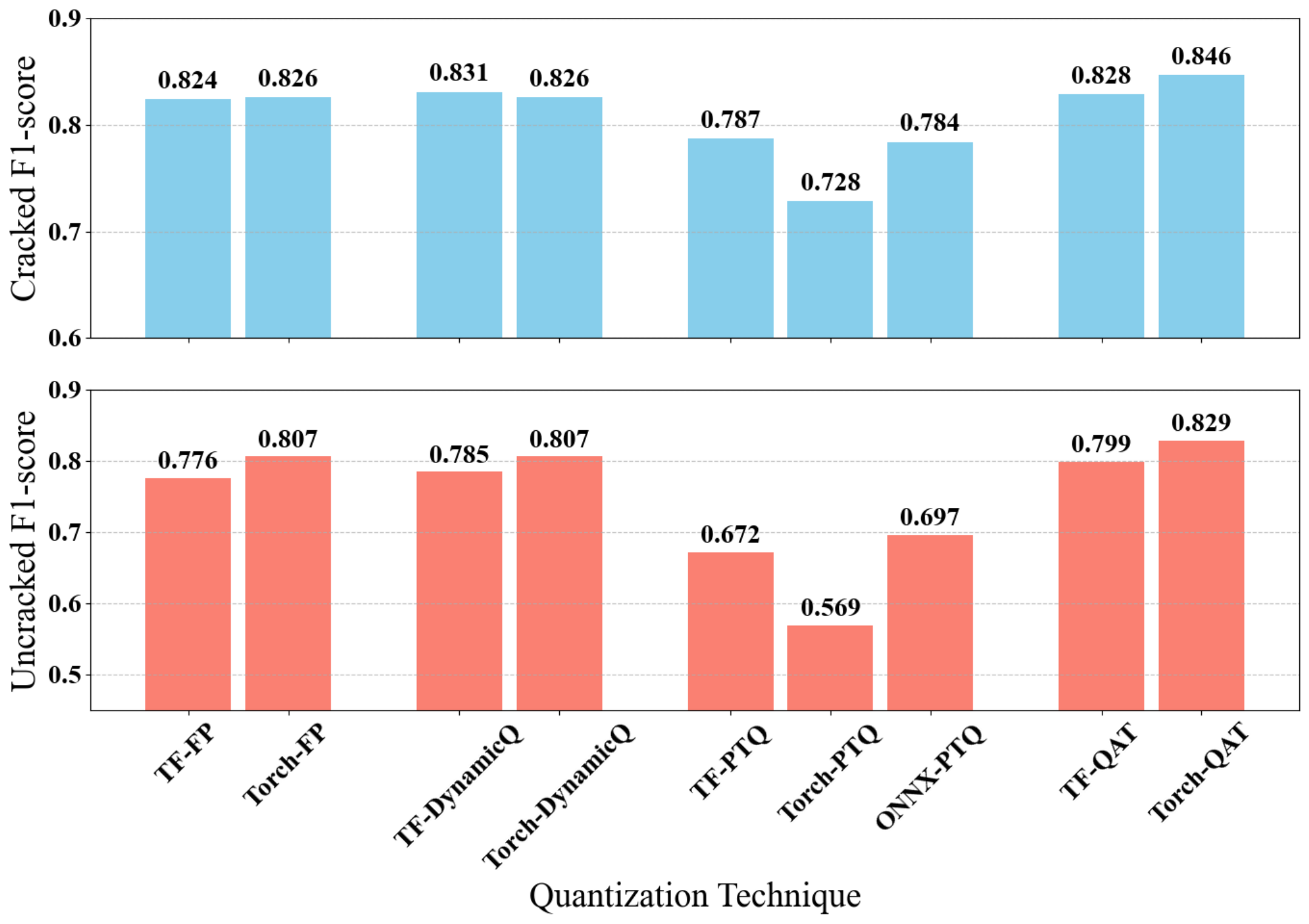}
      \caption{Performance of different quantization techniques with TF and Pytorch on MBNV2x0.5.}
      \label{fig:moibilenetv2}
\end{figure}

\subsection{On-device Performance}

%Power = 1.12W
\begin{table*}[t]
\caption{Deployed Model Performances on STM32H7B3I-DK Discovery Kit.}
\begin{center}
\label{tab:model_deployed_performances}
\begin{tabular}{lcccccc}
\toprule
\makecell[c]{\textbf{\textit{Quantization techniques}}\\\textbf{\textit{with Models \& Frameworks}}} & \textbf{\textit{F1-score}} & \makecell[c]{\textbf{\textit{Flash}}\\\textbf{\textit{(Bytes)}}} & \makecell[c]{\textbf{\textit{RAM}}\\\textbf{\textit{(Bytes)}}} & \makecell[c]{\textbf{\textit{Inference Time}}\\\textbf{\textit{(ms)}}} & \makecell[c]{\textbf{\textit{MACC}}} & \makecell[c]{\textbf{\textit{Energy Consumption}}\\\textbf{\textit{per Inference (mJ)}}} \\
\midrule
FP-TF-MBNV1x0.25 & 0.8285 & 877,608 & 833,432& 17.433 $\pm$ 0.199& 43,311,440 & 19.524\\
FP-TF-MBNV2x0.5 & 0.8001 & 2,773,820 & 2,487,808& - & 102,704,888 & - \\
\cmidrule(lr){1-7}
FP-Torch-MBNV1x0.25 & \textbf{0.8482} & 876,848& 1,220,468& 20.673 $\pm$ 0.234
& 42,126,002 & 23.153\\
FP-Torch-MBNV2x0.5 & 0.8161 & 2,762,460 & 2,488,392 & - & 97,818,882& - \\
\cmidrule(lr){1-7}
DynamicQ-TF-MBNV1x0.25 & 0.8214 & 877,596 & 833,432 & 17.310 $\pm$ 0.243& 43,311,440 & 19.387\\
DynamicQ-TF-MBNV2x0.5 & 0.8079 & 2,773,804 & 2,487,808 & - & 102,704,888 & - \\
\cmidrule(lr){1-7}
PTQ-TF-MBNV1x0.25 & 0.6843 & 291,945& 620,992& \textbf{14.776 $\pm$ 0.069}& \textbf{41,091,156} & \textbf{16.549}\\
PTQ-TF-MBNV2x0.5 & 0.7295 & 835,208& \textbf{655,108}& 35.588 $\pm$ 0.332& 96,348,220 & 39.858\\
\cmidrule(lr){1-7}
PTQ-Torch-MBNV1x0.25 & 0.7953 & 282,114 & 317,900 & 15.270 $\pm$ 0.151& 41,166,902 & 17.102\\
PTQ-Torch-MBNV2x0.5 & 0.6486 & 831,774 & 738,980 & 37.769 $\pm$ 0.259& 98,338,886 & 42.301\\
\cmidrule(lr){1-7}
PTQ-ONNX-MBNV1x0.25 & 0.7682 & \textbf{274,379} & \textbf{317,120} & 15.885 $\pm$ 0.123& 42,126,514 & 17.791\\
PTQ-ONNX-MBNV2x0.5 & 0.7403 & \textbf{786,829}& 732,864 & \textbf{34.832 $\pm$ 0.433}& \textbf{94,533,922}& \textbf{39.011}\\
\cmidrule(lr){1-7}
QAT-TF-MBNV1x0.25 & 0.8206 & 292,567 & 620,992 & 15.899 $\pm$ 0.128& 43,612,500 & 17.806\\
QAT-TF-MBNV2x0.5 & 0.8137 & 835,974 & 655,108& 39.471 $\pm$ 0.642& 103,005,948 & 44.207\\
\cmidrule(lr){1-7}
QAT-Torch-MBNV1x0.25 & 0.8114 & 282,114 & 317,900 & 15.298 $\pm$ 0.219& 41,166,902 & 17.133\\
QAT-Torch-MBNV2x0.5 & \textbf{0.8376}& 831,774 & 738,980 & 38.081 $\pm$ 0.779& 98,338,886 & 42.650\\
\cmidrule(lr){1-7}
DynamicQ-Torch-MBNV1x0.25 & 0.8482 & - & - & - & - & - \\
DynamicQ-Torch-MBNV2x0.5 & 0.8161 & - & - & - & - & - \\

\bottomrule
\end{tabular}
\end{center}
\end{table*}

This section analyzes the on-device performance of various quantization techniques, focusing on metrics such as accuracy (F1-score), memory usage (flash and RAM), inference time, MACC, and energy consumption per inference. The results in Table \ref{tab:model_deployed_performances} highlight the trade-offs between model accuracy and resource efficiency for vision-based crack detection tasks. Notably, the inference was run ten times and the average and uncertainty range values were calculated.

The FP models generally achieve the highest F1-scores. For instance, the FP-Torch-MBNV1x0.25 model achieves the best accuracy, with an F1-score of 0.8482. However, these models are resource-intensive, requiring significantly more flash memory, RAM, and energy. In contrast, dynamic quantization techniques result in a slight decrease in accuracy (e.g., DynamicQ-TF-MBNV1x0.25 achieves an F1-score of 0.8214). However, on the TF platform, there is minimal difference in RAM, flash, and MACC before and after quantization. Visualization using the Netron\footnote{https://netron.app/} revealed that the structure of \texttt{tflite} files of FP-TF-MBNV1x0.25 and DynamicQ-TF-MBNV1x0.25 are identical, with weights still in FP32 format, suggesting that the conversion to \texttt{tflite} or the quantization process may not have been successful. For PyTorch models, dynamic quantization could not be deployed, as ONNX does not support converting dynamically quantized models.

Compared to the FP models, PTQ significantly reduces flash and RAM usage. Taking MBNV1x0.25 as an example, PTQ-TF achieves the lowest inference time (14.776 ms), MACC (41,091,156), and energy consumption (16.549 mJ), but only reaches an F1 score of 0.6843. In contrast, PTQ-ONNX demonstrates the best resource efficiency, requiring only 274,379 bytes of flash and 317,120 bytes of RAM while achieving an F1 score of 0.7682. PTQ-Torch lies between the two in terms of memory and inference time, mainly due to structural changes introduced during the model conversion process. 
In PTQ-TF, the weights of each layer are stored in INT8 format, with the quantize operation added as a separate layer immediately after the input layer, and the dequantize operation added before the output classification layer. All computations are carried out in INT8 format. This technique results in the fastest inference time, lowest MACC, and minimal energy consumption. However, the high degree of quantization leads to the most significant accuracy degradation.

For PTQ-ONNX, weights are stored in INT8 format separately from each operation layer, while quantization parameters (mapping FP32 to INT8) are stored as FP values. Actual computation requires converting the stored INT8 weights back to FP32 for convolution calculations. Additionally, the outputs of each layer and the inputs to the next FP computation layer undergo \texttt{quantize} and \texttt{dequantize} operations. This additional processing leads to the highest MACC (42,126,514) but also achieves relatively high accuracy asncompared to PTQ-TF (F1-score: 0.7682). Flash and RAM usage is consistent across TF and ONNX platforms for MBNV1x0.25, but discrepancies arise for MBNV2x0.5. This difference is because MBNV1x0.25 uses ReLU as the activation function on all platforms, whereas MBNV2x0.5 uses ReLU in \texttt{tflite} but Clip in \texttt{onnx}.

For PTQ-Torch, weights are similarly stored separately, consistent with PTQ-ONNX quantization. Additionally, although the inputs to each layer has the same \texttt{quantize} operation, the subsequent \texttt{dequantize} operations differ. This technique introduces an additional Identity module to store parameters. This explains why PTQ-Torch-MBNV1x0.25 achieves the highest accuracy among the three PTQ processes but also consumes the most RAM. For MBNV2x0.5, the accuracy is the lowest, likely due to additional operations causing information loss in its unique inverted residual structure.

QAT, implemented in both TF and PyTorch frameworks, achieves a better balance between accuracy and efficiency. For example, QAT-TF-MBNV1x0.25 achieves an F1 score of 0.8206, significantly higher than PTQ, while maintaining nearly identical model structures, computational flows, and memory usage as PTQ-TF-MBNV1x0.25. The accuracy improvement is attributed to the inclusion of quantization during training. However, the downside of QAT is the need to retrain the model, which is more time-consuming and less convenient than PTQ.

In summary, PTQ deliver the best performance in inference time and energy consumption, making them the preferred choice for real-time applications, provided there is a sufficiently accurate model. On the other hand, QAT achieves accuracy comparable to FP models but requires additional training time, avoiding the accuracy degradation caused by PTQ. Dynamic quantization was in this study found to not be deployable successfully in any of the platforms.

\section{Conclusion and Future Work}

This study surveyed the performance of MBNV1x0.25 and MBNV2x0.5 models utilizing various quantization techniques for an on-device crack detection task the first time. The results show that while FP models achieved the highest F1-scores, they are resource-intensive and costly for deployment on constrained devices. PTQ provided significant resource savings, such as PTQ-ONNX reducing 68.74\% of flash usage and 61.95\% of RAM usage for MBNV1x0.25, but it leads to considerable accuracy degradation, with TF-PTQ scoring as low as 0.6424 for the non-crack task. 

In contrast, QAT consistently delivered FP-like accuracy (e.g., 0.8285 and 0.8463 for MBNV2x0.5 in the crack task using TF-QAT and Torch-QAT, respectively), while maintaining reasonable resource efficiency. Despite requiring additional training, QAT effectively mitigates the accuracy loss seen in PTQ, making it the most robust quantization technique for applications where accuracy is critical.

The results of this study highlight the importance of quantization techniques, particularly QAT, in deploying real-time, low-power vision-based crack detection systems for SHM. By enabling UAV-based autonomous inspections, this work enhances the safety, scalability, and cost-effectiveness of SHM applications.

Future research will focus on improving PTQ to enhance accuracy, addressing deployment challenges of dynamic quantization, and optimizing TinyML models for specific hardware. Additionally, validating QAT-based models in real-world UAV-supported SHM scenarios, considering environmental and operational factors, will be prioritized to better support autonomous SHM inspections.

\section*{Acknowledgement}
The authors would like to thank the financial support by the Knowledge Foundation under grant 20180170 (NIIT) and 20240029-H-02 (TransTech2Horizon).

\balance
\bibliographystyle{./IEEEtran}
\bibliography{IEEEi2mtc2025}

\begin{thebibliography}{10}
\providecommand{\url}[1]{#1}
\csname url@samestyle\endcsname
\providecommand{\newblock}{\relax}
\providecommand{\bibinfo}[2]{#2}
\providecommand{\BIBentrySTDinterwordspacing}{\spaceskip=0pt\relax}
\providecommand{\BIBentryALTinterwordstretchfactor}{4}
\providecommand{\BIBentryALTinterwordspacing}{\spaceskip=\fontdimen2\font plus
\BIBentryALTinterwordstretchfactor\fontdimen3\font minus \fontdimen4\font\relax}
\providecommand{\BIBforeignlanguage}[2]{{%
\expandafter\ifx\csname l@#1\endcsname\relax
\typeout{** WARNING: IEEEtran.bst: No hyphenation pattern has been}%
\typeout{** loaded for the language `#1'. Using the pattern for}%
\typeout{** the default language instead.}%
\else
\language=\csname l@#1\endcsname
\fi
#2}}
\providecommand{\BIBdecl}{\relax}
\BIBdecl

\bibitem{adin2023b}
V.~Ad{\i}n, Y.~Zhang, B.~Oelmann, and S.~Bader, ``Tiny {{Machine Learning}} for {{Damage Classification}} in {{Concrete Using Acoustic Emission Signals}},'' in \emph{2023 {{IEEE International Instrumentation}} and {{Measurement Technology Conference}} ({{I2MTC}})}, May 2023, pp. 1--6.

\bibitem{huang2024b}
C.~Huang, X.~Sun, and Y.~Zhang, ``Tiny-{{Machine-Learning-Based Supply Canal Surface Condition Monitoring}},'' \emph{Sensors}, vol.~24, no.~13, p. 4124, Jan. 2024.

\bibitem{9335288}
H.~Zhao, G.~Chen, H.~Hong, and X.~Zhu, ``Remote structural health monitoring for industrial wind turbines using short-range doppler radar,'' \emph{IEEE Transactions on Instrumentation and Measurement}, vol.~70, pp. 1--9, 2021.

\bibitem{zhangtim2023a}
Y.~Zhang, V.~Adin, S.~Bader, and B.~Oelmann, ``Leveraging {{Acoustic Emission}} and {{Machine Learning}} for {{Concrete Materials Damage Classification}} on {{Embedded Devices}},'' \emph{IEEE Transactions on Instrumentation and Measurement}, vol.~72, pp. 1--8, 2023.

\bibitem{9964291}
S.~Song, X.~Zhang, Y.~Chang, and Y.~Shen, ``An improved structural health monitoring method utilizing sparse representation for acoustic emission signals in rails,'' \emph{IEEE Transactions on Instrumentation and Measurement}, vol.~72, pp. 1--11, 2023.

\bibitem{Ud2024}
U.~Muthumala, Y.~Zhang, L.~S. Martinez-Rau, and S.~Bader, ``Comparison of tiny machine learning techniques for embedded acoustic emission analysis,'' in \emph{2024 IEEE 10th World Forum on Internet of Things (WF-IoT)}, 2024.

\bibitem{7991459}
S.~Dorafshan, M.~Maguire, N.~V. Hoffer, and C.~Coopmans, ``Challenges in bridge inspection using small unmanned aerial systems: Results and lessons learned,'' in \emph{2017 International Conference on Unmanned Aircraft Systems (ICUAS)}, 2017, pp. 1722--1730.

\bibitem{10156415}
J.~Heichel, R.~Mitra, F.~Jafari, A.~Das, S.~Dorafshan, and N.~Kaabouch, ``A system for real-time display and interactive training of predictive structural defect models deployed on uav,'' in \emph{2023 International Conference on Unmanned Aircraft Systems (ICUAS)}, 2023, pp. 1221--1225.

\bibitem{10568977}
K.-W. Tse, R.~Pi, W.~Yang, X.~Yu, and C.-Y. Wen, ``Advancing uav-based inspection system: The ussa-net segmentation approach to crack quantification,'' \emph{IEEE Transactions on Instrumentation and Measurement}, vol.~73, pp. 1--14, 2024.

\bibitem{8453409}
S.~Dorafshan, R.~J. Thomas, C.~Coopmans, and M.~Maguire, ``Deep learning neural networks for suas-assisted structural inspections: Feasibility and application,'' in \emph{2018 International Conference on Unmanned Aircraft Systems (ICUAS)}, 2018, pp. 874--882.

\bibitem{zakariaBridge2022a}
M.~Zakaria, E.~Karaaslan, and F.~N. Catbas, ``Advanced bridge visual inspection using real-time machine learning in edge devices,'' \emph{Advances in Bridge Engineering}, vol.~3, no.~1, p.~27, Dec. 2022.

\bibitem{Michele2023I2MTC}
S.~Heo, N.~Baumann, C.~Margelisch, M.~Giordano, and M.~Magno, ``Low-cost smart raven deterrent system with tiny machine learning for smart agriculture,'' in \emph{2023 IEEE International Instrumentation and Measurement Technology Conference (I2MTC)}, 2023, pp. 1--6.

\bibitem{LucSAS2024}
L.~S. Martinez-Rau, Y.~Zhang, B.~Oelmann, and S.~Bader, ``Tinyml anomaly detection for industrial machines with periodic duty cycles,'' in \emph{2024 IEEE Sensors Applications Symposium (SAS)}, 2024, pp. 1--6.

\bibitem{Michele2022TIM1}
M.~Giordano, N.~Baumann, M.~Crabolu, R.~Fischer, G.~Bellusci, and M.~Magno, ``Design and performance evaluation of an ultralow-power smart iot device with embedded tinyml for asset activity monitoring,'' \emph{IEEE Transactions on Instrumentation and Measurement}, vol.~71, pp. 1--11, 2022.

\bibitem{Michele2022TIM2}
X.~Wang, F.~Geiger, V.~Niculescu, M.~Magno, and L.~Benini, ``Leveraging tactile sensors for low latency embedded smart hands for prosthetic and robotic applications,'' \emph{IEEE Transactions on Instrumentation and Measurement}, vol.~71, pp. 1--14, 2022.

\bibitem{zhangsas2024b}
Y.~Zhang, L.~S. {Martinez-Rau}, B.~Oelmann, and S.~Bader, ``Enabling {{Autonomous Structural Inspections}} with {{Tiny Machine Learning}} on {{UAVs}},'' in \emph{2024 {{IEEE Sensors Applications Symposium}} ({{SAS}})}, Jul. 2024, pp. 1--6.

\bibitem{YuxuanZh2025}
Y.~Zhang, L.~S. Martinez-Rau, B.~Oelmann, and S.~Bader, ``On-device autonomous vision-based crack detection using tinyml,'' \emph{IEEE Transactions on Instrumentation and Measurement}, 2024, under-review.

\bibitem{9869850}
H.-H. Chin, R.-S. Tsay, and H.-I. Wu, ``An adaptive high-performance quantization approach for resource-constrained cnn inference,'' in \emph{2022 IEEE 4th International Conference on Artificial Intelligence Circuits and Systems (AICAS)}, 2022, pp. 336--339.

\bibitem{dorafshanSDNET2018AnnotatedImage2018}
S.~Dorafshan, R.~J. Thomas, and M.~Maguire, ``{{SDNET2018}}: {{An}} annotated image dataset for non-contact concrete crack detection using deep convolutional neural networks,'' \emph{Data in Brief}, vol.~21, pp. 1664--1668, Dec. 2018.

\end{thebibliography}
\vspace{12pt}
\end{document}